\documentclass[10pt,twocolumn,letterpaper]{article}

\usepackage{cvpr}              

\usepackage{amsmath,amsfonts,bm}

\def\eqref#1{equation~\ref{#1}}

\def\1{\bm{1}}

\DeclareMathAlphabet{\mathsfit}{\encodingdefault}{\sfdefault}{m}{sl}
\SetMathAlphabet{\mathsfit}{bold}{\encodingdefault}{\sfdefault}{bx}{n}

\usepackage[utf8]{inputenc} 
\usepackage[T1]{fontenc}    
\usepackage{url}            
\usepackage{booktabs}       
\usepackage{amsfonts}       
\usepackage{nicefrac}       
\usepackage{microtype}      
\usepackage{graphicx}
\usepackage{amsmath,amssymb,amsthm}
\usepackage{multirow}
\usepackage{makecell}
\usepackage{url}

\usepackage{caption}
\usepackage{subcaption}
\usepackage{tablefootnote}

\usepackage[accsupp]{axessibility}

\usepackage[utf8]{inputenc}
\usepackage[english]{babel}
\usepackage[dvipsnames]{xcolor}
\usepackage[tableposition=top]{caption}

\definecolor{myblue}{RGB}{0, 128, 255}
\definecolor{mycolor}{RGB}{147,112,219}

\definecolor{citecolor}{HTML}{0071bc}

\definecolor{citecolor}{HTML}{0071bc}

\newcommand{\proposedregmethod}{\textit{Listen2Student}}

\usepackage[pagebackref,breaklinks,colorlinks, citecolor=citecolor]{hyperref}

\usepackage[capitalize]{cleveref}
\crefname{section}{Sec.}{Secs.}
\Crefname{section}{Section}{Sections}
\Crefname{table}{Table}{Tables}
\crefname{table}{Tab.}{Tabs.}

\def\cvprPaperID{1554} 
\def\confName{CVPR}
\def\confYear{2022}

\begin{document}

\title{
Unbiased Teacher v2: Semi-supervised Object Detection\\for Anchor-free and Anchor-based Detectors}

%

\author{%
   Yen-Cheng Liu$^{1}$, Chih-Yao Ma$^{2}$, Zsolt Kira$^{1}$
   
   \\  $^{1}$Georgia Institute of Technology, $^{2}$Meta\\ \texttt{\{ycliu, zkira\}@gatech.edu}, \texttt{cyma@fb.com} \\
}

\maketitle

\begin{abstract}
With the recent development of Semi-Supervised Object Detection (SS-OD) techniques, object detectors can be improved by using a limited amount of labeled data and abundant unlabeled data. 
However, there are still two challenges that are not addressed:
(1) there is no prior SS-OD work on anchor-free detectors, and (2) prior works are ineffective when pseudo-labeling bounding box regression.
In this paper, we present Unbiased Teacher v2, which shows the generalization of SS-OD method to anchor-free detectors and also introduces \proposedregmethod\ mechanism for the unsupervised regression loss.
Specifically, we first present a study examining the effectiveness of existing SS-OD methods on anchor-free detectors and find that they achieve much lower performance improvements under the semi-supervised setting.
We also observe that box selection with centerness and the localization-based labeling used in anchor-free detectors cannot work well under the semi-supervised setting. 
On the other hand, our \proposedregmethod\ mechanism explicitly prevents misleading pseudo-labels in the training of bounding box regression; we specifically develop a novel pseudo-labeling selection mechanism based on the Teacher and Student's \textit{relative} uncertainties.
This idea contributes to favorable improvement in the regression branch in the semi-supervised setting.
Our method, which works for both anchor-free and anchor-based methods, consistently performs favorably against the state-of-the-art methods in \textit{VOC}, \textit{COCO-standard}, and \textit{COCO-additional}.
\end{abstract}


\section{Introduction}
\label{sec:intro}
Deep learning models have achieved remarkable performance on object detection tasks in recent years, though the strong performance heavily relies on training a network with abundant images with human-annotated labels. 
To reduce the label supervision for training object detectors, Semi-Supervised Object Detection (SS-OD) methods have been proposed to leverage only limited labeled data but more abundant unlabeled data to improve performance~\cite{jeong2019consistency,gao2019note,sohn2020simple,liu2021unbiased,zhou2021instant}.
Existing state-of-the-art SS-OD methods apply self-training techniques, which generate pseudo-labels and enforce the consistency between unlabeled data with different augmentations. 
Despite the significant improvement, 
there are still two remaining issues that are left untackled:
(1) \textbf{there is no prior SS-OD work on anchor-free detectors} and (2) \textbf{prior works are ineffective in pseudo-labeling on the bounding box regression}.

\begin{figure*}[t]
   \begin{picture}(0,200)
     \put(0,10){\includegraphics[width=\linewidth]{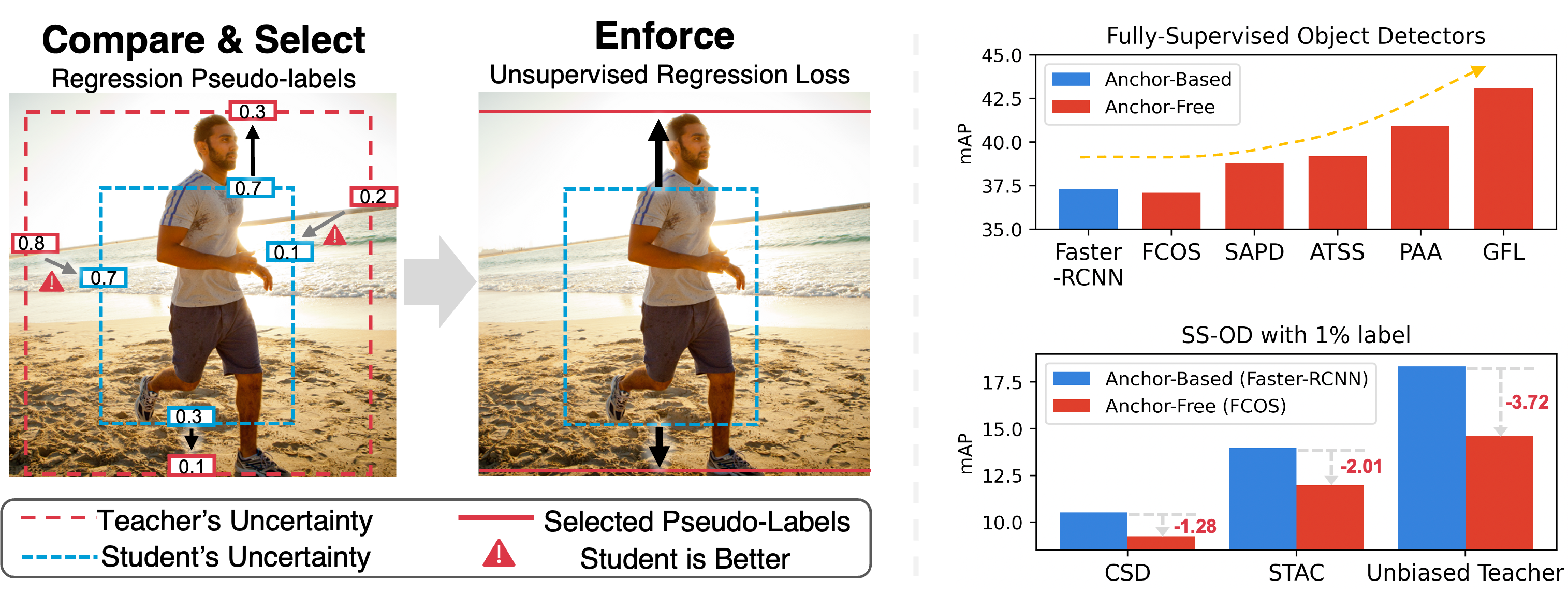}}
     \put(136,0){(a)}
     \put(405,0){(b)}
   \end{picture}
    \caption{  
    To improve the unsupervised regression loss, we propose (a) \proposedregmethod, which explicitly compares the prediction uncertainties between the Teacher and the Student and selects these instances where the  teacher has lower uncertainty than the student.
    We then enforce the unsupervised regression loss on these selected regression pseudo-labels.  (b) Anchor-free detectors are rapidly developed recently, while adapting the pseudo-labeling method on the anchor-free models results in less improvements compared with the anchor-based detectors.
    }
    \label{fig:teaser}
    \vspace{-5mm}
\end{figure*}

First, anchor-free detectors have been recently getting more attention in the community of object detection~\cite{tian2019fcos,zhang2020bridging,zhu2019feature,wang2019region,kong2020foveabox,li2020generalized,li2020gflv2}, with the promise of achieving  competitive accuracy, computational efficiency, and potential generalization to new datasets or environments~\cite{zhang2020bridging}. 
In spite of these advances,
existing SS-OD works~\cite{jeong2019consistency, sohn2020simple, liu2021unbiased} mainly focus on anchor-based detectors (\textit{e.g.,} Faster-RCNN~\cite{ren2015faster} and SSD~\cite{liu2016ssd}) but do not empirically verify their effectiveness on anchor-free detectors. 
In fact, when we adapt recent state-of-the-art SS-OD methods to anchor-free detectors, we observe that, compared with its improvement on anchor-based models, the improvement is much smaller on anchor-free models (see Figure~\ref{fig:teaser}\textcolor{red}{a} and Table~\ref{table:anchor2free}). 
With extensive analysis provided in Section~\ref{sec:pseudo-label_on_anchor_free}, we find that \textit{some advanced techniques performing favorably in the fully-supervised setting do not work in the semi-supervised setting with limited supervision}. 
For example, the centerness score becomes unreliable for box selection under the  semi-supervised setting, and the localization-based labeling method is not robust to the localization noise in pseudo-labels.

Second, following the Teacher-Student framework, the existing SS-OD works~\cite{sohn2020simple,zhou2021instant} apply an unsupervised regression loss with the pseudo-boxes generated from confidence thresholding (\textit{i.e.,} a threshold on the box score).
However, we find that this approach inherits some potential issues that can be further addressed.  
For instance, (1) instead of using one single metric (\textit{e.g.,} box score or box IoU) to \textit{jointly} represent the quality of four boundaries, the confidence/uncertainty of \textbf{each} boundary should be predicted individually;
(2) confidence in the  classification branch might not be able to reflect the quality of boundary prediction on the regression branch. Instead, we propose to predict uncertainties on the regression branch to select pseudo-labels for boundary prediction;
(3) Lastly, simply relying on Teacher's confidence/uncertainty prediction to select pseudo labels for regression cannot prevent \textit{misleading} instances for the regression task. Instead, we propose to exploit the \textit{relative uncertainties between the Teacher and Student} to select the boundary-level pseudo-labels, in which the Teacher has lower uncertainty than the Student.
Integrating the three components, we propose \proposedregmethod\ to improve the unsupervised regression loss for the SS-OD tasks, as shown in Figure~\ref{fig:teaser}b.

We demonstrate that our proposed method achieves significant improvements compared to the state-of-the-art SS-OD methods when using \textit{both} anchor-free and anchor-based detectors on several SS-OD benchmarks, including \textit{COCO-standard}, \textit{COCO-additional}, and \textit{VOC}. 
We also provide ablation studies to examine the effectiveness of our~\proposedregmethod.
We summarize the main contributions as follows:

\begin{itemize}
    \item We show the generalization of our proposed semi-supervised method on both anchor-based and anchor-free detectors. To the best of our knowledge, we are the first to examine the anchor-free models on SS-OD, and we identify core issues in applying SS-OD methods on anchor-free detectors.
    \item We explicitly remove misleading instances in regression pseudo-labels by considering \textit{relative} uncertainty estimation from the Teacher and Student predictions. We provide analyses to verify effectiveness of our approach on anchor-free and anchor-based detectors.
    \item Based on our empirical study on anchor-free and anchor-based detectors, our method shows favorable improvements against the state-of-the-art methods. With the proposed method, we also bridge the performance gap between anchor-free and anchor-based detectors under the semi-supervised setting.
    \end{itemize}

\section{Related Work}
\label{sec:related}

\textbf{Anchor-Free Object Detectors.}
The development of deep learning models has resulted in significant improvements on object detection tasks. 
Existing object detectors consist of anchor-based detectors~\cite{ren2015faster,liu2016ssd,cai2016unified,shrivastava2016training,lin2017focal,xie2017aggregated} and anchor-free detectors~\cite{law2018cornernet,zhou2019bottom,tian2019fcos,zhu2019feature,wang2019region,kong2020foveabox}.
Specifically, anchor-based detectors predict the box shift and scaling for the pre-defined anchor-boxes, and each predicted box is labeled according to its intersection-over-union (IoU) score to the ground-truth boxes. 
Based on label assignment (\textit{i.e.,} assign classification labels to predicted instances)
and sub-sampling of foreground-background anchor boxes, the models are then trained to perform object detection.
Despite remarkable results that have been achieved, applying anchor-based detectors on new datasets requires experts to tune hyper-parameters~\cite{jiao2019survey} related to anchor-boxes, which limits the ability to adapt to new datasets or environments~\cite{zhang2020bridging}.

Alternatively, anchor-free models alleviate these concerns by removing the pre-defined anchor-boxes in detection models.
For example, keypoint-based anchor-free detectors eliminated the need for designing a set of anchor boxes by representing a box as two corner points~\cite{law2018cornernet}, a center point with four extreme points~\cite{zhou2019bottom}, and a center point with the box weight and height~\cite{zhou2019objects}.
Similarly, FCOS~\cite{tian2019fcos} removed the pre-defined anchor-boxes and predicted a classification score, distances to four boundaries, and a centerness score for each pixel.
Several works improved the performance of the anchor-free model by proposing an adaptive sample selection~\cite{zhang2020bridging}, jointly training the centerness and classification branches with soft-labels~\cite{li2020generalized}, soft-selecting the pyramid levels~\cite{zhu2019soft}, and modeling the boundary uncertainty~\cite{li2020gflv2}. 
In this paper, we use FCOS~\cite{tian2019fcos} as our base anchor-free model, since it is publicly available and widely used in existing anchor-free models~\cite{zhang2020bridging,zhu2019soft,li2020generalized,li2020gflv2}.

\textbf{Semi-Supervised Object Detection.}
Semi-supervised learning (SSL) for image classification has been rapidly developed and obtained promising results in recent years. Existing SSL image classification works~\cite{berthelot2019mixmatch,laine2016temporal,sajjadi2016regularization,tarvainen2017mean,zhang2018mixup,yun2019cutmix,guo2019mixup,hendrycks2020augmix,sohn2020fixmatch} apply input augmentations/perturbations and consistency regularization on unlabeled images to improve the model trained with the limited amount of labeled data.
Inspired by these works, several semi-supervised object detection works have been proposed to exploit similar ideas to train object detectors in a semi-supervised manner.
For example, CSD~\cite{jeong2019consistency} apply a left-right consistency loss to enforce prediction consistency between horizontally flipped unlabeled images.
Some other works~\cite{sohn2020simple,liu2021unbiased,zhou2021instant,tang2021humble} exploit pseudo-labeling, where a model iteratively generates the pseudo-labels of unlabeled data and add the confident predictions into the training data.
STAC~\cite{sohn2020simple} uses the limited amount of labeled data to train an object detector, which is used to generate the pseudo-labels for unlabeled data in an offline manner. 
To refine the quality of pseudo-labels, Instant-Teaching~\cite{zhou2021instant} proposes a co-rectify scheme to rectify the false prediction between two identical but independently trained models. 
Humble Teacher~\cite{tang2021humble} applies exponential moving average (EMA) and soft pseudo-labels to improve against the model trained on labeled data only.
Unbiased Teacher~\cite{liu2021unbiased} proposes to generate the pseudo-labels in an online fashion, and the quality of pseudo-labels is further improved by addressing the pseudo-labeling bias issue.
SoftTeacher~\cite{xu2021end} proposes a simple background-weighted loss and box variance filter to improve performance against the supervised baselines. 
While they can improve the performance in the semi-supervised setting, existing works only present their results on anchor-based detectors.
We are thus interested in investigating the generalization of the state-of-the-art methods (\textit{i.e.,} pseudo-labeling) on anchor-free models and improving the performance of anchor-free models for semi-supervised object detection tasks.

\section{Method}

\subsection{Background: Semi-supervised Object Detection and Pseudo-labeling}
\label{sec:problem_def}

With the goal of learning an object detector in a semi-supervised setting, we assume a set of labeled images $\bm{D}_s =\{\bm{x}^s_i, \bm{y}^s_i\}^{N_s}_{i=1}$ and  unlabeled images $\bm{D}_u=\{\bm{x}^u_i\}^{N_u}_{i=1}$ are available during training.

In order to address semi-supervised object detection, existing works~\cite{sohn2020simple, zhou2021instant, liu2021unbiased} exploit the pseudo-labeling method. 
Specifically, this line of works contains two stages: 1) The burn-in stage and 2) the mutual learning stage. 
In the burn-in stage, with the available labeled data, an initial object detector is trained with the standard supervised losses, $\mathcal{L}_{sup} = \sum_i \mathcal{L}(\bm{x}^s_i, \bm{y}^s_i)$.
In the mutual learning stage, the pretrained object detector is duplicated into a Student and a Teacher model initially. 
Then, in each training iteration, the Teacher model takes the weakly-augmented unlabeled images as input and predicts the bounding boxes, and the instances with the box score higher than a threshold $\tau$ (\textit{i.e.,} confidence thresholding) are selected as the pseudo-labels.

Based on the pseudo-labels and the same unlabeled image but with a stronger augmentation, the unsupervised loss $\mathcal{L}_{unsup}$ is computed and combined with the supervised loss $\mathcal{L}_{sup}$ to train the Student model, $\theta_s \leftarrow \theta_s + \gamma \frac{\partial (\mathcal{L}_{sup} + \bm{\lambda}_u \mathcal{L}_{unsup}) }{\partial\theta_s}$, where $\mathcal{L}_{unsup} =  \sum_i     \mathcal{L}(\bm{x}^u_i, \bm{\hat{y}}^u_i)$. 
To refine the quality of the pseudo-labels, the Teacher model weight ($\theta_t$) can be further updated with the Student model weight ($\theta_s$) via Exponential Moving Average (EMA) as shown in \cite{liu2021unbiased}.

Although the existing works~\cite{sohn2020simple,zhou2021instant,liu2021unbiased} based on pseudo-labeling have presented significant improvements on anchor-based detectors (\textit{i.e.,} Faster-RCNN), it is still unclear whether such a method is applicable to anchor-free detectors. 
This motivates us to investigate its generalization to anchor-free detectors, and we provide our findings and show that the state-of-the-art SS-OD method is not effective as it is mostly designed for anchor-based detectors (in Section~\ref{sec:pseudo-label_on_anchor_free}).
\begin{table}[t]
\centering
\renewcommand{\arraystretch}{1.3}
\caption{
    \textbf{Adaption of Unbiased Teacher~\cite{liu2021unbiased} to an anchor-free model}. The performance is degraded when applying Unbiased Teacher on the anchor-free model (FCOS). 
    }
\label{table:anchor2free}
\resizebox{\linewidth}{!}{%
\begin{tabular}{llcccc|c}
\toprule
           & &\multicolumn{4}{c|}{COCO-standard}   \\ \cline{3-7}
    Methods       & Models &\multicolumn{1}{c}{0.5\%}  &\multicolumn{1}{c}{1\%}     & \multicolumn{1}{c}{5\%}    & \multicolumn{1}{c}{10\%}  & \multicolumn{1}{|c}{100\%} \\ \hline
UT~\cite{liu2021unbiased} &  F-RCNN  & 14.36 &   18.33  &  26.65 &  29.56 &  37.90  \\ 
UT~\cite{liu2021unbiased} &  FCOS         & 10.27  \scriptsize{\textcolor{red}{(-4.09)}}&   14.61 \scriptsize{\textcolor{red}{(-3.72)}}  &  23.99 \scriptsize{\textcolor{red}{(-2.66)}} & 28.18 \scriptsize{\textcolor{red}{(-1.38)}} & 38.10 \\
\bottomrule
\end{tabular}
}
\vspace{-3mm}
\end{table}


\subsection{Pseudo-labeling on Anchor-Free Detectors} 
\label{sec:pseudo-label_on_anchor_free}

\begin{figure}[t]
   \begin{picture}(0,90)
     \put(0,5){\includegraphics[width=\linewidth]{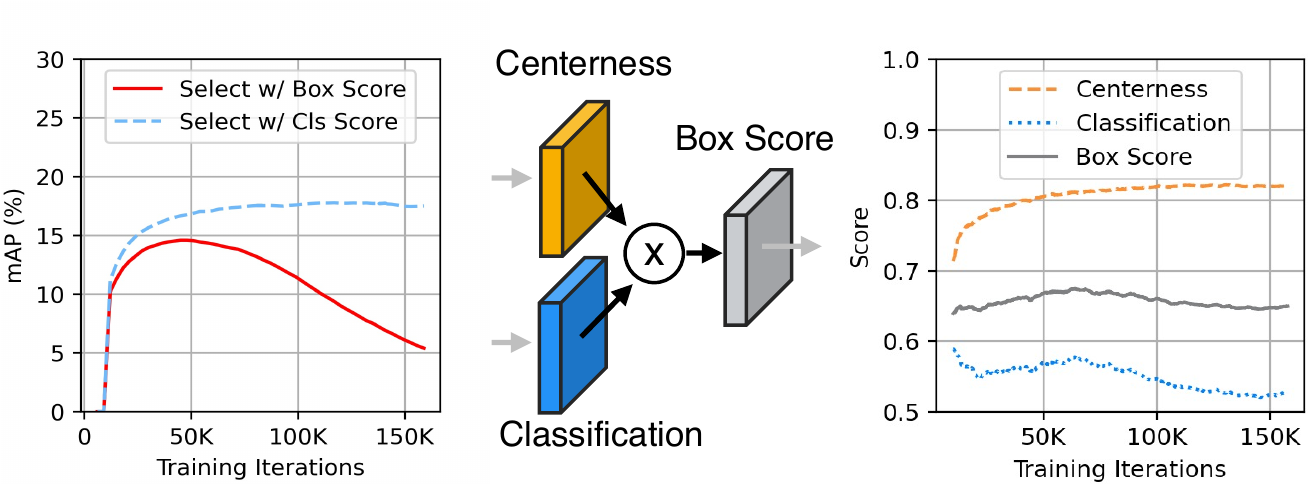}}
     \put(29,-3){(a)}
     \put(108,-3){(b)}
     \put(193,-3){(c)}

   \end{picture}
    \caption{
    \textbf{Illustration of Centerness bias issue.} 
    \textbf{(a)} Selecting pseudo-boxes based on box scores leads to worse results in semi-supervised learning compared with selecting based on classification scores. 
    \textbf{(b)} Box scores of the anchor-free detectors~\cite{tian2019fcos,zhang2020bridging} are defined as the multiplication of the centerness scores and the classification scores, and we find that \textbf{(c)} box scores of pseudo-boxes is dominated by the centerness scores, which are unreliable in semi-supervised setting (see Appendix for further details). }
    \label{fig:centerness_bias}
\end{figure}
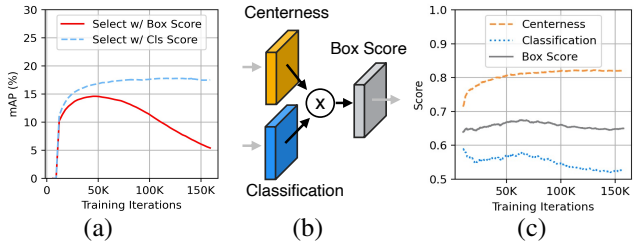

We take the widely used FCOS model~\cite{tian2019fcos} as an example of anchor-free detector for studying SS-OD task.
FCOS~\cite{tian2019fcos} has three major prediction branches: 1) a classifier for performing object category classification, 2) a centerness branch for indicating the probability of being the center of foreground objects, and 3) a regressor for estimating the distances to the boundaries of an object. 
These models usually exploit fully convolutional layers and perform pixel-wise predictions.
To train the model, all pixels inside the ground-truth boxes are labeled as foreground and the remaining pixels as background, and regression loss and centerness loss are only enforced in these foreground instances. 
For more details of anchor-free detectors, please refer to the FCOS paper~\cite{tian2019fcos}.

As shown in Figure~\ref{fig:teaser}b and Table~\ref{table:anchor2free}, we observe that simply applying the existing state-of-the-art SS-OD methods~\cite{jeong2019consistency, sohn2020simple, liu2021unbiased} on anchor-free detectors obtains much smaller improvements compared with anchor-based detectors.
We attribute this to the following two factors.
\begin{table}[t]
\centering
\renewcommand{\arraystretch}{1.3}
\caption{
While box selection based on box score leads to higher detection accuracy in fully-supervised settings, it performs worse than box selection based on classification scores under semi-supervised learning settings.
Fully-supervised results are from FCOS~\cite{tian2019fcos}.}
\label{table:pseudo-label-select}
\resizebox{0.8\linewidth}{!}{%
\begin{tabular}{cccc}
\toprule
                           & Class. score & Box score & $\Delta$ \\ \specialrule{1pt}{1pt}{1pt}
Fully-supervised           & 33.50 & 37.10 &  \textbf{\textcolor{myblue}{+3.60}}  \\ 
Semi-supervised            & 17.79 & 15.12 &  \textbf{\textcolor{red}{-2.67}}     \\ 
\bottomrule
\end{tabular}
}
\end{table}

\noindent\textbf{Centerness bias issue.} 
As presented in Figure~\ref{fig:centerness_bias}\textcolor{red}{b} and Table~\ref{table:pseudo-label-select}, we notice that selecting the pseudo-boxes based on box scores performs worse than solely relying on classification scores in the semi-supervised setting, while FCOS~\cite{tian2019fcos} shows using box scores leads to better results in the fully-supervised setting. 
We observed that this is because the box scores of some anchor-free detectors~\cite{tian2019fcos,zhang2020bridging} are defined as the multiplication of classification scores and centerness scores (see Figure~\ref{fig:centerness_bias}\textcolor{red}{a}), and the pseudo-boxes selected based on the box scores have relatively high centerness scores but low classification scores (see Figure~\ref{fig:centerness_bias}\textcolor{red}{c}). 
This reveals that the box scores are dominated by the centerness scores in the pseudo-labeling mechanism. 
However, with the limited amount of labels used in the training, the centerness scores are not reliable for reflecting whether a prediction is a foreground instance since there is no supervision to suppress the centerness scores for background instances in the centerness branch\footnote{A similar observation was also made in Generalized Focal Loss~\cite{li2020generalized}.}. 
As a result, these selected high centerness pseudo-boxes are likely to be the background instances, and adding these false-positive pseudo-boxes in the semi-supervised training degrades the effectiveness of the pseudo-labeling and also aggravates the centerness bias issue.

\begin{figure}[t]
\begin{picture}(0,90)
 \put(0,5){\includegraphics[width=\linewidth]{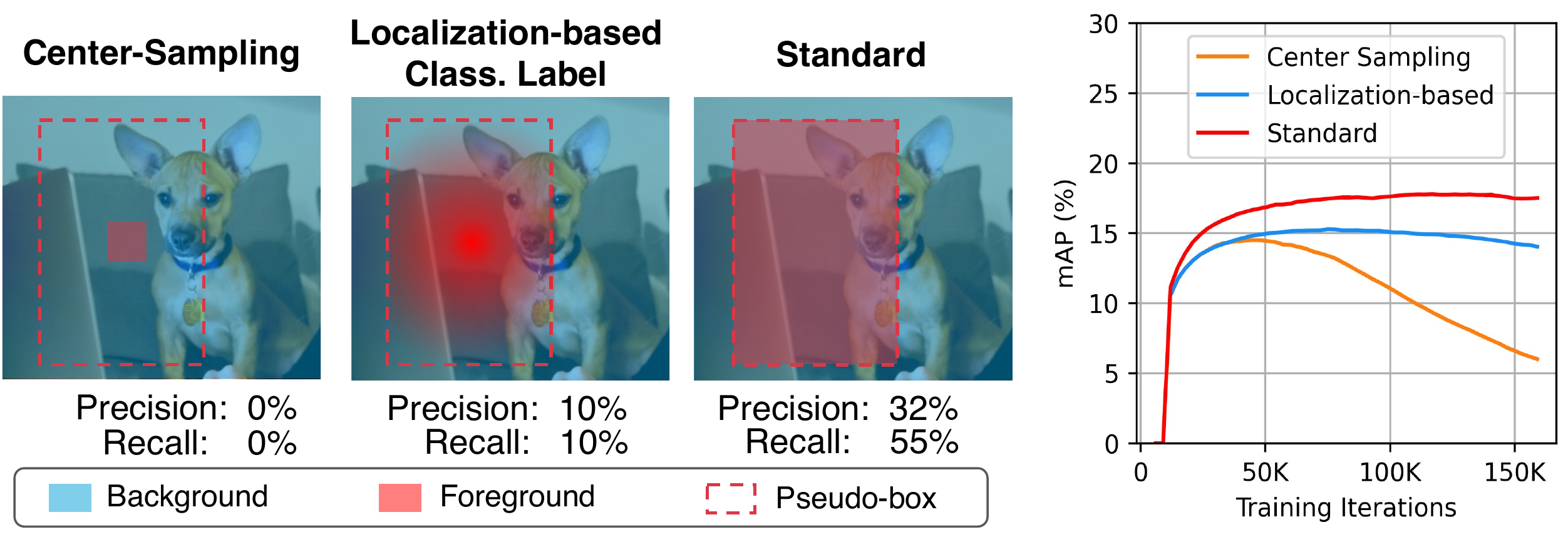}}
 \put(75,-3){(a)}
 \put(200,-3){(b)}

\end{picture}
\caption{\textbf{Illustration of Unreliable label assignments.}
(a) Existing techniques for improving fully-supervised anchor-free detectors such as Center Sampling~\cite{tian2019fcos} and localization-based classification labels~\cite{li2020generalized} are less robust to the localization noise (\textit{e.g.,} box center shifted) in pseudo-boxes, and the pixel-wise recall and precision of these two techniques are lower than the standard label assignment. Thus, (b) our empirical evaluation shows that the standard label assignment leads to a better result.
} 
\label{fig:label_assignment_vis}
\end{figure}

\noindent\textbf{Unreliable Label Assignment.}
\begin{table}[t]
\centering
\renewcommand{\arraystretch}{1.3}
\caption{
While the Center-Sampling improves the anchor-free detectors in fully-supervised setting, it degrades the performance in the semi-supervised setting. Fully-supervised results from FCOS~\cite{tian2019fcos}. 
}
\label{table:center_sampling}
\resizebox{\linewidth}{!}{%
\begin{tabular}{cccc}
\toprule
                           & w/o Center-Sampling & w/ Center-Sampling & $\Delta$ \\ \specialrule{1pt}{1pt}{1pt}
Fully-supervised           & 37.10 & 38.10 &  \textbf{\textcolor{myblue}{+1.00}}  \\ 
Semi-supervised            & 17.79 & 14.96 &  \textbf{\textcolor{red}{-2.83}}     \\ 
\bottomrule
\end{tabular}
}
\end{table}

To improve the performance of the \textit{fully-supervised} anchor-free detector, several works~\cite{li2020generalized,zhou2019objects} proposed to use soft classification labels, which are weighted based on the bounding box localization as presented in Figure~\ref{fig:label_assignment_vis}\textcolor{red}{a}.
Similarly, FCOS~\cite{tian2019fcos} also presented an advanced label assignment technique, \textit{center-sampling}, which regards the instances close to the center of the object as foreground instances and improves against the model using the standard label assignment that labels all instances insides ground-truth boxes as foreground and the remaining instances as background.
\textit{Although the above techniques improve the anchor-free detectors during fully-supervised training, we found that they are not effective or even detrimental during semi-supervised training} (see Figure~\ref{fig:label_assignment_vis}\textcolor{red}{b} and Table~\ref{table:center_sampling}).  
We hypothesize that this is because the pseudo-boxes can have localization noise (either due to the center of the box being shifted or the box has incorrect width and height), and using \textit{center-sampling} or the \textit{Localization-based soft labels} makes pixel-wise predictions incorrectly labeled as either foreground (false positive) or background (false negative).
For instance, as shown in Figure~\ref{fig:label_assignment_vis}, the precision and recall of \textit{center-sampling} is much lower than \textit{standard} for this particular example with reasonable amount of localization noise.


\begin{figure}[t]
   \begin{picture}(0,90)
     \put(0,0){\includegraphics[width=\linewidth]{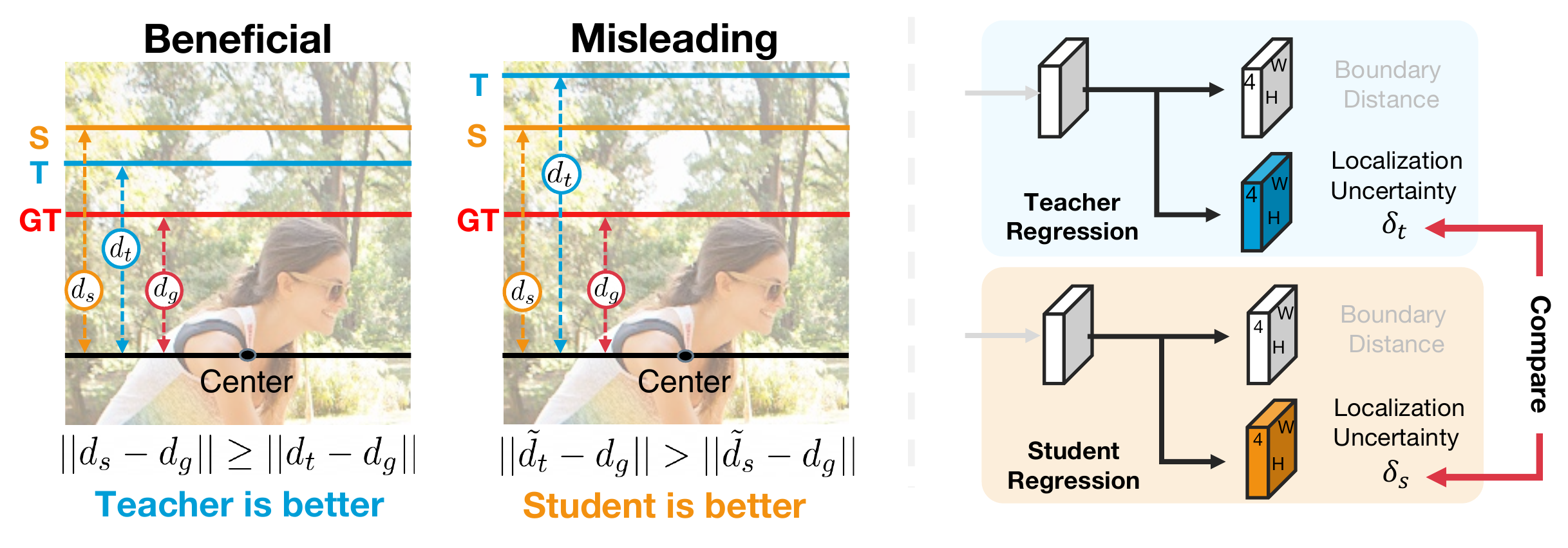}}
     \put(65,-5){(a)}
     \put(180,-5){(b)}
   \end{picture}
    \caption{ \textbf{(a) Beneficial/Misleading regression pseudo-labels  and (b) \proposedregmethod.} (a) We categorize the regression pseudo-labels into beneficial and misleading instances, and (b) our \proposedregmethod\ prevents the misleading regression pseudo-labels and thus improves the regression branch. 
    }
    \label{fig:benefiting_misleading}
\end{figure}

To alleviate the centerness bias issue, we select the pseudo-boxes based on classification score only (and ignore the centerness score) in limited-supervision scenarios, since we empirically found that classification score is more reliable to represent the objectness of the predicted instances especially under limited supervision.
In this way, the false-positive pseudo-labels are less likely to impede the effectiveness of pseudo-labeling and thus improve the performance of pseudo-labeling. 
We also train the classifier with the hard labels (\textit{i.e.,} one-hot vector) rather than the soft labels with the box localization weighting. 
Finally, instead of using the \textit{center-sampling}, we use the \textit{standard} label assignment method, which labels all elements inside the bounding boxes as the foreground and the remaining as the background. 

\subsection{\proposedregmethod\ for Unsupervised Regression Loss}
\label{sec:unsup_reg}

\subsubsection{Limitations of confidence thresholding for regression}
While confidence thresholding has been demonstrated to work well in classification (image-level~\cite{sohn2020fixmatch} or box-level~\cite{sohn2020simple,liu2021unbiased,zhou2021instant}), we observed solely relying on the box confidence cannot effectively remove the \textit{misleading} instances in the box regression, and there are several reasons why it does not perform favorably: 
(1) First, the confidence thresholding in existing works selects pseudo-boxes based on the box scores, which only reflect the confidence of object classification in Faster-RCNN~\cite{liu2021unbiased}, and there is no explicit module estimating the confidence (or uncertainty) of regression prediction, \textit{i.e.,} the regression branch only predicts the boundary location without any metric indicating localization uncertainty in vanilla object detectors.
(2) Second, using one single score (\textit{e.g.,} centerness or IoU score) to jointly represent the quality of four predicted boundaries is not accurate, as it is hard to obtain a pseudo-box with four \textbf{equally precise} boundaries under the limited-supervision setting. 
(3) Lastly, unlike pseudo-labels for discrete object categories, the real-valued regression outputs are unbounded.
Selecting pseudo-boxes solely based on the Teacher's confidence cannot explicitly prevent \textit{misleading} instances in the pseudo label for regression, because the Teacher can still provide a regression direction that is contradictory to the ground-truth direction. Similar observations are also found in prior works in knowledge distillation for regression tasks~\cite{chen2017learning,saputra2019distilling}.

\subsubsection{\proposedregmethod}
To address the above concerns and improve the regression branch with the Teacher-Student mechanism,  we aim to select the \textit{beneficial} instances and remove \textit{misleading} instances for the training of the regression branch.
Intuitively, we develop a novel way to use \textit{relative} prediction information between the Student and Teacher; to our knowledge this is the first instance of moving beyond using just the Teacher's prediction information.
Specifically, as shown in Figure~\ref{fig:benefiting_misleading}, the \textit{beneficial} instance of boundary prediction is defined as: the instance that satisfies $|| \tilde{d}_t - d_g || \leq || \tilde{d}_s - d_g||$,
where $\tilde{d}_t$ is the Teacher's regression prediction, $\tilde{d}_s$ is the Student's regression prediction, and $d_g$ is the ground-truth regression label. 
As a comparison, the \textit{misleading} instance of regression is expressed as the instance satisfying $|| \tilde{d}_t - d_g || > || \tilde{d}_s - d_g||$.

\textbf{Uncertainty prediction for regression.}
While we were hoping to use ground-truth labels $d_g$ to decide whether the predictions from the Teacher is better or not, in reality, the ground-truth labels are not available for SS-OD.
Therefore, we propose to predict the localization uncertainty, which loosely correlates with the error to the ground-truth label (\textit{i.e.,} $||\tilde{d}_t-d_g||$ and $||\tilde{d}_s-d_g||$) for the unlabeled data.
As shown in Figure~\ref{fig:benefiting_misleading}, the localization uncertainty of each boundary prediction is derived by adding an additional branch, which has the same output size as the boundary distance regression branch.
The localization uncertainty branch is jointly trained with the boundary distance branch, and we use the negative power log-likelihood loss (NPLL)~\cite{lee2020localization}\footnote{~\proposedregmethod\ is not limited to NPLL, and other regression uncertainty estimation methods~\cite{he2019bounding} are also potentially applicable.} as the regression loss, 
\begin{align*}
\quad\mathcal{L}^{sup}_{reg} =  \sum_i \eta_i (\sum (\frac{{(d_s - d_g)}^2}{2 {\delta_s}^2}+ \frac{1}{2} \log {\delta_s}^2) +2 \log 2\pi),
 \end{align*}
where $\eta_i$ is the IoU score between the predicted box and the ground-truth box, and $\delta_s$ is the predicted uncertainty of the Student. 

\textbf{Relative uncertainties for pseudo-label selection.}
With the uncertainty estimation, we first loosely remove the boundaries where student has very small localization uncertainty $\delta_s\leq\sigma_s$.
We then propose a selection mechanism which explicitly takes not only the  Teacher's localization uncertainty $\delta^i_t$ but also the Student's localization uncertainty $\delta^i_s$ into account for the pseudo-label selection.
By selecting the \textit{beneficial} instances where the Teacher has lower localization uncertainty than the Student with a margin $\sigma$, our unsupervised regression loss is thus defined as
\begin{equation}
\mathcal{L}^{unsup}_{reg}=\begin{cases}
    \sum_i ||\tilde{d}^i_t - \tilde{d}^i_s ||, & \text{if $\delta^i_t + \sigma \leq \delta^i_s$ }\\
    0, & \text{otherwise}
  \end{cases},
\end{equation}
where $\sigma \geq 0$ is a margin between the localization uncertainties of Teacher and Student. 
Note that the unsupervised regression loss is computed in the boundary level rather than the box level, so some boundaries of a box are used to computed unsupervised regression loss while the others are not. 

The core idea of this mechanism is that the Teacher should only guide the Student with the instances that the Teacher has lower uncertainty than the student, as it indicates that the Teacher has a potentially lower error.
By contrast, for the instances, which the Teacher has higher uncertainty than the Student, we should not enforce the loss, as the Teacher is likely to predict worse than the Student and thus mislead the Student for these instances.
Based on this selection mechanism, we can explicitly prevent gradients from \textit{misleading} instances from degrading the performance of the regression branch. 
Our regression branch can thus gradually be refined and obtain more accurate boundary prediction.
It is worth noting that the localization uncertainty branch is an individual branch and only used in the training stage, thus introducing no additional computation during inference. 

\section{Experiments}\label{sec:expt}
\begin{table*}[t]
\centering
\renewcommand{\arraystretch}{1.3}
\caption{
    {Experimental results of the anchor-free model (FCOS-ResNet50) on \textit{COCO-standard}}. * We reimplement and adapt to FCOS-ResNet50. 
    We randomly sample labeled data and run each method $5$ times, and we report the mean and standard deviation for each result.
    We used $8$ labeled images and $8$ unlabeled images for all results presented in this table. 
    }
\label{table:mscoco_standard_anchor_free}
\resizebox{\linewidth}{!}{%
\begin{tabular}{llllll}
\toprule
           & \multicolumn{5}{c}{\textbf{Anchor-free detectors} on COCO-standard}  \\ \cline{2-6}
           & \multicolumn{1}{c}{0.5\%}
  &\multicolumn{1}{c}{1\%}     & \multicolumn{1}{c}{2\%}    & \multicolumn{1}{c}{5\%}    & \multicolumn{1}{c}{10\%}     \\ \hline
Supervised&  5.42 $\pm$ 0.01 &  8.43 $\pm$ 0.03    &    11.97 $\pm$ 0.03 &    17.01 $\pm$  0.01 &  20.98  $\pm$ 0.01        \\ \hline
CSD~\cite{jeong2019consistency}*     & 5.76 $\pm$ 0.55  \scriptsize{\textcolor{myblue}{(+0.34)}} & 9.23 $\pm$ 0.08  \scriptsize{\textcolor{myblue}{(+0.80)}} & 12.53 $\pm$ 0.04  \scriptsize{\textcolor{myblue}{(+0.56)}} & 18.09 $\pm$ 0.08 \scriptsize{\textcolor{myblue}{(+1.08)}} & 22.06 $\pm$ 0.01 \scriptsize{\textcolor{myblue}{(+1.08)}} \\ 
STAC~\cite{sohn2020simple}*     & 8.79 $\pm$ 0.12 \scriptsize{\textcolor{myblue}{(+3.37)}} &   11.97 $\pm$ 0.12 \scriptsize{\textcolor{myblue}{(+3.54)}} & 15.50 $\pm$ 0.16 \scriptsize{\textcolor{myblue}{(+3.53)}}&  20.36 $\pm$ 0.05 \scriptsize{\textcolor{myblue}{(+3.35)}} &  24.31 $\pm$ 0.02 \scriptsize{\textcolor{myblue}{(+3.33)}}\\ 
Unbiased Teacher~\cite{liu2021unbiased}*     & 10.27  $\pm$ 0.13 \scriptsize{\textcolor{myblue}{(+4.85)}} &  14.61 $\pm$ 0.10 \scriptsize{\textcolor{myblue}{(+6.18)}} & 18.70 $\pm$ 0.21 \scriptsize{\textcolor{myblue}{(+6.73)}} &  23.99 $\pm$ 0.12 \scriptsize{\textcolor{myblue}{(+6.98)}} &  28.18 $\pm$ 0.01 \scriptsize{\textcolor{myblue}{(+7.20)}} \\ 
\textbf{Ours}      & \textbf{16.25 $\pm$ 0.18 \scriptsize{\textcolor{myblue}{(+10.83)}}}& \textbf{22.71 $\pm$ 0.42  \scriptsize{\textcolor{myblue}{(+14.28)}} }& \textbf{26.03 $\pm$ 0.12 \scriptsize{\textcolor{myblue}{(+14.06)}}} &  \textbf{30.08 $\pm$ 0.04 \scriptsize{\textcolor{myblue}{(+13.07)}}}  &  \textbf{32.61 $\pm$ 0.03 \scriptsize{\textcolor{myblue}{(+11.63)}}} \\
\bottomrule
\end{tabular}
}
\end{table*}

\begin{table*}[t]
\centering
\renewcommand{\arraystretch}{1.3}
\caption{
    Experimental results of anchor-based models (FasterRCNN-ResNet50) on \textit{COCO-standard}. For a fair comparison, we make the batch size consistent to the baseline methods. $\dagger$: using labeled/unlabeled batch size $32/32$, *: using batch size labeled/unlabeled batch size $8/40$, and rest of the results using batch size $8/8$. We randomly sample labeled data and run each method $5$ times, and we report the mean and standard deviation for each result. 
    }
\label{table:mscoco_standard_fasterrcnn}
\resizebox{\linewidth}{!}{%
\begin{tabular}{cccccc}
\toprule
           & \multicolumn{5}{c}{{\textbf{Anchor-based detectors} on COCO-standard}}  \\ \cline{2-6}
           & \multicolumn{1}{c}{0.5\%}
  &\multicolumn{1}{c}{1\%}     & \multicolumn{1}{c}{2\%}    & \multicolumn{1}{c}{5\%}    & \multicolumn{1}{c}{10\%}     \\ \hline
Supervised& 6.83 $\pm$ 0.15 & 9.05 $\pm$ 0.16      &    12.70 $\pm$ 0.15    &    18.47 $\pm$ 0.22    &    23.86 $\pm$ 0.81          \\ \hline
CSD~\cite{jeong2019consistency}     & 7.41 $\pm$ 0.21 \scriptsize{\textcolor{myblue}{(+0.58)}}& 10.51 $\pm$ 0.06  \scriptsize{\textcolor{myblue}{(+1.46)}}   & 13.93 $\pm$ 0.12 \scriptsize{\textcolor{myblue}{(+1.23)}} & 18.63 $\pm$ 0.07  \scriptsize{\textcolor{myblue}{(+0.16)}} & 22.46 $\pm$ 0.08  \scriptsize{\textcolor{red}{(-1.40)}}  \\ 
STAC~\cite{sohn2020simple}     & 9.78 $\pm$ 0.53 \scriptsize{\textcolor{myblue}{(+2.95)}} & 13.97 $\pm$ 0.35 \scriptsize{\textcolor{myblue}{(+4.92)}}   & 18.25 $\pm$ 0.25 \scriptsize{\textcolor{myblue}{(+5.55)}} & 24.38 $\pm$ 0.12 \scriptsize{\textcolor{myblue}{(+5.86)}} & 28.64 $\pm$ 0.21 \scriptsize{\textcolor{myblue}{(+4.78)}}  \\ 
Humble Teacher~\cite{tang2021humble}     & - & 16.96 $\pm$ 0.38 \scriptsize{\textcolor{myblue}{(+7.91)}}   & 21.72 $\pm$ 0.24 \scriptsize{\textcolor{myblue}{(+9.02)}} & 27.70 $\pm$ 0.15 \scriptsize{\textcolor{myblue}{(+9.23)}} & 31.61 $\pm$ 0.28 \scriptsize{\textcolor{myblue}{(+7.74)}}  \\ 
Instant Teaching~\cite{zhou2021instant}     & - & 18.05 $\pm$ 0.15 \scriptsize{\textcolor{myblue}{(+9.00)}}   & 22.45 $\pm$ 0.15 \scriptsize{\textcolor{myblue}{(+9.75)}} & 26.75 $\pm$ 0.05 \scriptsize{\textcolor{myblue}{(+8.28)}} & 30.40 $\pm$ 0.05 \scriptsize{\textcolor{myblue}{(+6.54)}}  \\ 
Unbiased Teacher~\cite{liu2021unbiased} & 14.36 $\pm$ 0.09 \scriptsize{\textcolor{myblue}{(+7.53)}} & 18.33 $\pm$ 0.19 \scriptsize{\textcolor{myblue}{(+9.28)}}   & 22.23 $\pm$ 0.21 \scriptsize{\textcolor{myblue}{(+9.53)}} & 26.65 $\pm$ 0.31 \scriptsize{\textcolor{myblue}{(+8.18)}} & 29.56 $\pm$ 0.24 \scriptsize{\textcolor{myblue}{(+5.70)}}  \\ 
ISMT~\cite{yang2021interactive}     & - & 18.88 $\pm$ 0.74 \scriptsize{\textcolor{myblue}{(+9.83)}}   & 22.43 $\pm$ 0.56 \scriptsize{\textcolor{myblue}{(+9.73)}} & 26.37 $\pm$ 0.24 \scriptsize{\textcolor{myblue}{(+7.90)}} & 30.53 $\pm$ 0.52 \scriptsize{\textcolor{myblue}{(+6.67)}}  \\
\textbf{Ours}    &  \textbf{17.51 $\pm$ 0.24 \scriptsize{\textcolor{myblue}{(+10.68)}}}& \textbf{21.84  $\pm$ 0.13 \scriptsize{\textcolor{myblue}{(+12.79)}}} & \textbf{ 26.14  $\pm$ 0.01 \scriptsize{\textcolor{myblue}{(+13.44)}}}  & \textbf{30.06   $\pm$ 0.14 \scriptsize{\textcolor{myblue}{(+11.59)}}}    & \textbf{33.50  $\pm$ 0.03 \scriptsize{\textcolor{myblue}{(+9.64)}}}      \\

\hline
SoftTeacher~\cite{xu2021end}$*$    &  -  & 20.46 $\pm$ 0.39 \scriptsize{\textcolor{myblue}{(+11.41)}} & - & 30.74 $\pm$ 0.08   \scriptsize{\textcolor{myblue}{(+12.27)}}  & 34.04 $\pm$ 0.14 \scriptsize{\textcolor{myblue}{(+10.18)}}   \\ 
\textbf{Ours}*    &  \textbf{21.02 $\pm$ 0.49 \scriptsize{\textcolor{myblue}{(+14.19)}}}& \textbf{24.79  $\pm$ 0.30 \scriptsize{\textcolor{myblue}{(+15.74)}}} & \textbf{ 28.23  $\pm$ 0.05 \scriptsize{\textcolor{myblue}{(+15.53)}}}  & \textbf{32.05   $\pm$ 0.04 \scriptsize{\textcolor{myblue}{(+13.58)}}}    & \textbf{35.02  $\pm$ 0.02 \scriptsize{\textcolor{myblue}{(+11.16)}}}      \\

\hline
Unbiased Teacher~\cite{liu2021unbiased}$\dagger$    &  16.94 $\pm$ 0.23 \scriptsize{\textcolor{myblue}{(+10.11)}}  & 20.75 $\pm$ 0.12 \scriptsize{\textcolor{myblue}{(+11.72)}}  & 24.30 $\pm$ 0.07   \scriptsize{\textcolor{myblue}{(+11.60)}}  & 28.27 $\pm$ 0.11 \scriptsize{\textcolor{myblue}{(+9.80)}}   & 31.50 $\pm$ 0.10 \scriptsize{\textcolor{myblue}{(+7.64)}}         \\
\textbf{Ours$\dagger$}    &  \textbf{21.26 $\pm$ 0.21 \scriptsize{\textcolor{myblue}{(+14.43)}}}& \textbf{25.40  $\pm$ 0.36 \scriptsize{\textcolor{myblue}{(+16.35)}}} & \textbf{ 28.37  $\pm$ 0.03 \scriptsize{\textcolor{myblue}{(+15.67)}}}  & \textbf{31.85   $\pm$ 0.09 \scriptsize{\textcolor{myblue}{(+13.38)}}}    & \textbf{  35.08 $\pm$ 0.02 \scriptsize{\textcolor{myblue}{(+11.22)}}}      \\
\bottomrule
\end{tabular}}
\end{table*}

\begin{table*}[h]
\centering
\renewcommand{\arraystretch}{1.3}
\caption{Average precision (AP) breakdown of unsupervised regression methods. We also report the absolute improvement of each unsupervised regression loss method against the model without the unsupervised regression loss.}
\label{table:ap_breakdown_all}
\resizebox{0.75\linewidth}{!}{%
\begin{tabular}{ccccccccccc}
\toprule
                        &    & AP55  & AP60  & AP65  & AP 70  &  AP75& AP80 & AP85 & AP90 & AP95\\ \midrule
No regression &   & 29.71 & 27.34 & 24.64 &  21.38 & 17.55 & 13.27 & 8.33  & 3.45 & 0.35  \\ \midrule
\multirow{2}{*}{Confidence Thresholding} &    & 30.60 & 28.19 & 25.07 &  21.93 & 17.96 & 13.32 & 8.22 & 3.12 & 0.32 \\ 
&    & \textcolor{myblue}{+0.89} & \textcolor{myblue}{+0.85} & \textcolor{myblue}{+0.43} &  \textcolor{myblue}{+0.55} & \textcolor{myblue}{+0.41} & \textcolor{myblue}{+0.05} & \textcolor{red}{-0.11} & \textcolor{red}{-0.33} & \textcolor{red}{-0.03} \\ \midrule
\multirow{2}{*}{\proposedregmethod\ (Ours)} &   & 30.78 & 28.59 & 26.19 &  23.05 & 19.64 & 15.61 & 10.47 & 5.06 & 0.58 \\ 
&  &  \textcolor{myblue}{+1.07} & \textcolor{myblue}{+1.25} & \textcolor{myblue}{+1.56} &  \textcolor{myblue}{+1.67} & \textcolor{myblue}{+2.09} & \textcolor{myblue}{+2.34} & \textcolor{myblue}{+2.14} & \textcolor{myblue}{+1.61} & \textcolor{myblue}{+0.23} \\ 
\bottomrule
\end{tabular}
}
\end{table*}

\begin{table}[t]
\centering
\renewcommand{\arraystretch}{1.3}
\caption{
    Results of the \textbf{Anchor-based model (Faster-RCNN)} on \textit{VOC}.  
    }
\vspace{-3mm}
\label{table:voc}
\resizebox{\linewidth}{!}{%
\begin{tabular}{ccccc}
\toprule
Methods          & Labeled                  & Unlabeled              & $AP^{50}$  & $AP^{50:95}$ \\\midrule
Supervised       & \multirow{7}{*}{VOC07}   & \multirow{7}{*}{VOC12} & 76.70 & 43.60   \\
STAC~\cite{sohn2020simple}  &                          &             & 77.45 & 44.64   \\
ISMT~\cite{yang2021interactive}             &     &                  & 77.23 & 46.23   \\
Instant-Teaching~\cite{zhou2021instant} &                          &                & 79.20 & 50.00   \\
Humble Teacher~\cite{tang2021humble}   &                          &                        & 80.94 & 53.04   \\
Unbiased Teacher~\cite{liu2021unbiased} &                          &                        & 80.51 & 54.48   \\
Ours             &                          &                        & \textbf{81.29} & \textbf{56.87}   \\\midrule 
STAC~\cite{sohn2020simple}             &   \multirow{6}{*}{VOC07} &   \multirow{6}{*}{\begin{tabular}[c]{@{}c@{}}VOC12 \\+\\ \textit{COCO20cls}\end{tabular}}   & 79.08 & 46.01   \\
ISMT~\cite{yang2021interactive}              &                          &                        & 77.75 & 49.59   \\
Instant-Teaching~\cite{zhou2021instant} &                          &                        & 79.00 & 50.80.  \\
Humble Teacher~\cite{tang2021humble}   &                          &                        & 81.29 & 54.41   \\
Unbiased Teacher~\cite{liu2021unbiased} &                          &                        & 81.71 & 55.79	   \\
Ours             &                          &                        & \textbf{82.04} & \textbf{58.08}   \\ 
\bottomrule
\end{tabular}
}
\vspace{-1mm}
\end{table}

\subsection{Settings and Implementation Details}

\textbf{Experimental Settings.} \label{sec:expt_setting}
We follow the experimental settings presented in the existing semi-supervised object detection works~\cite{sohn2020simple,liu2021unbiased}.
Specifically, we use MS-COCO~\cite{lin2014microsoft} and PASCAL VOC~\cite{everingham2010pascal} and examine our proposed method on three experimental scenarios, \textit{COCO-standard}, \textit{COCO-additional}, and \textit{VOC}.
For \textit{COCO-standard}, we randomly sample $0.5$, $1$, $2$, $5$, and $10\%$ labeled training data as our labeled set, and the remaining data as the unlabeled set. 
For \textit{COCO-additional}, we use \textit{COCO2017-labeled} as labeled set and \textit{COCO2017-unlabeled} as the unlabeled set. 
We evaluate on \textit{COCO2017-val} for both \textit{COCO-standard} and \textit{COCO-additional} as  in previous works. 
As for \textit{VOC}, \textit{VOC2007-trainval} is used as the labeled set, and \textit{VOC2012-trainval} and \textit{COCO20cls} are used as the unlabeled set. 
All trained models in \textit{VOC} experiment are evaluated on \textit{VOC2007-test}.

\textbf{Model Architecture.} 
In order to examine the effectiveness of anchor-free models for semi-supervised object detection, we chose FCOS~\cite{tian2019fcos} as our base anchor-free models since it is widely adopted in existing anchor-free works~\cite{zhu2019soft,zhang2020bridging,li2020generalized,li2020gflv2}. 
As the existing works mainly focus on anchor-based models and use Faster-RCNN~\cite{sohn2020simple,jeong2019consistency,liu2021unbiased} or SSD~\cite{jeong2019consistency}, we also adapt the existing SS-OD methods~\cite{sohn2020simple,liu2021unbiased,jeong2019consistency} to the anchor-free model (\textit{e.g.,} FCOS).

\textbf{Implementation Details.}
Our implementation is based on Detectron2~\cite{wu2019detectron2}. 
To train our model, we use SGD optimizer with the learning rate $0.01$, and each batch contains $8$ labeled images and $8$ unlabeled images unless specified. 
We use the unsupervised loss weight $\lambda_{u}=3.0$ and classification threshold $\tau=0.5$, and we set $\sigma=0.1$ as the margin between localization uncertainties of Teacher and Student and $\sigma_s=0.5$.
We adapt the data augmentation used in Unbiased Teacher and applied the scale jittering used in SoftTeacher~\cite{xu2021end} without using any geometric augmentation during training, as we empirically find that the scale jittering leads to a significant improvement. 
More details are listed in the supplementary material.

\subsection{Results on Anchor-free Detector}
\textbf{COCO-standard.} 
We adapt three anchor-based methods, CSD~\cite{jeong2019consistency}, STAC~\cite{sohn2020simple}, and Unbiased Teacher~\cite{liu2021unbiased}, to the anchor-free models,
and each method was ran five times and their means and variances are reported, as presented in Table~\ref{table:mscoco_standard_anchor_free}.
Our model consistently performs favorably against the baseline methods under different degrees of supervision, and the improvement gap is larger when the level of supervision is lower. 
Our experiments on VOC and COCO-additional also result in a similar trend as well (see Appendix for experimental results).

\subsection{Results on Anchor-based Detector}
In addition to the results on the anchor-free model, we are also interested whether our proposed method can generalize to different types of object detectors.
Specifically, we apply our unsupervised regression loss on Unbiased Teacher and modify the regression branch to predict the localization uncertainty with an additional branch as we did in Section~\ref{sec:unsup_reg}. 
We examine our ~\proposedregmethod\ on the Faster-RCNN for \textit{COCO-standard}, \textit{VOC}, and \textit{COCO-additional} as follows.

\textbf{COCO-standard.}
As presented in Table~\ref{table:mscoco_standard_fasterrcnn}, compared with the state-of-the-art SS-OD methods~\cite{tang2021humble,liu2021unbiased,xu2021end}, our method obtains higher mAP under the cases where $0.5\%$ to $10\%$ data are labeled. 
Under different batch sizes, we could maintain the improvement gap against existing SS-OD methods and further improve the performance to $35.08$ mAP under \textit{COCO-standard} 10\% case. 
In addition, we also find that the performance gap between the anchor-free and anchor-based detectors is reduced by using our framework, and this verifies the generalization of our proposed~\proposedregmethod\ to both anchor-free and anchor-based detectors.

\textbf{VOC and COCO-additional.}
To verify whether our framework can improve the object detector trained with the unlabeled set, we also consider \textit{VOC} in Table~\ref{table:voc} and \textit{COCO-additional} in Table~\ref{table:coco_add_anchor}. 
With \textit{VOC07} used as the labeled set, our model can leverage \textit{VOC12} to achieve $56.87$ mAP, and using \textit{VOC12+COCO20cls} as the unlabeled set can further improve the model and achieve $58.08$mAP.
On the other hand, with the \textit{COCO2017-unlabeled} set, our model can perform favorably against the object detector trained on \textit{COCO2017-train} and achieve $44.75$ mAP. 
Note that we train our model for $720$k iterations and do not tune the inference threshold (same as SoftTeacher). Training the model longer or tuning the inference threshold can potentially further improve the performance.
These results confirm the effectiveness of our framework on improving the existing object detector using the extra unlabeled images. 

\subsection{Effectiveness of Unsupervised Regression Loss}
We compare the methods including 1) our proposed~\proposedregmethod\, 2) No unsupervised regression loss, and 3) using confidence thresholding and enforcing L1 loss, as used in existing works~\cite{sohn2020simple,zhou2021instant}. 
To further understand how these methods contribute to the improvement of bounding box regression, we provide an mAP breakdown from AP$55$ to AP$95$ of each method in Table~\ref{table:ap_breakdown_all}.
It is worth noting that we only change the unsupervised regression loss across these methods and keep the remaining objective functions and modifications the same across all variants. 

We observe that, although the confidence thresholding can improve the easier evaluation metrics (\textit{e.g.,} AP$55$), it cannot improve or even degrades the results on stricter evaluation metrics (\textit{e.g.,} AP$95$).
This shows that simply using the confidence thresholding cannot prevent misleading pseudo-labels from degrading the performance on extremely precise boundary predictions. 
In contrast, our \proposedregmethod\ shows consistent improvements on all evaluation metrics and leads to favorable results, especially on these stricter evaluation metrics.
This empirically confirms that our \proposedregmethod\ contributes to the more precise bounding box prediction, as our \proposedregmethod\ enforces the boundary-wise unsupervised regression loss, which exploits the pseudo-labels derived by comparing the uncertainty estimation of each boundary prediction. 

\textbf{Limitations and future works.}
Although we have shown the improvement and generalization on anchor-free and anchor-based detectors, applying SSOD methods on a large-scale unlabeled dataset (\textit{e.g.,} OpenImage) remains a challenge. 
We also find that the localization uncertainty estimation for boundary prediction leaves room for improvement to be integrated with the relative thresholding mechanism.  
There are other challenges such as unseen objects in the unlabeled dataset or domain shift between datasets. 
While these topics are not our focus in this paper, they are worth exploring in future research.

\begin{table}[t]
\centering
\renewcommand{\arraystretch}{1.3}
\caption{
    Results of the \textbf{Anchor-based model (Faster-RCNN)} on \textit{COCO-additional}. *We adapt the scale jitter used in SoftTeacher~\cite{xu2021end} to Unbiased Teacher and it leads to significant improvement.  
    }
\label{table:coco_add_anchor}
\resizebox{0.6\linewidth}{!}{%
\begin{tabular}{ccc}
\toprule
Methods                                 &      & $mAP$ \\\midrule
Supervised                              &      & 40.90   \\
CSD~\cite{jeong2019consistency}         &      & 38.52   \\
STAC~\cite{sohn2020simple}              &      & 39.21   \\
Humble Teacher~\cite{tang2021humble}    &      & 42.37   \\
Unbiased Teacher*~\cite{liu2021unbiased} &      & 44.06   \\
SoftTeacher~\cite{xu2021end}               &                    & 44.50 \\
Ours                                    &      & \textbf{44.75}   \\
\bottomrule
\end{tabular}
}
\vspace{-1mm}
\end{table}

\section{Conclusion}
In this paper, we examined the existing SS-OD methods on anchor-free models and presented the SS-OD benchmarks on anchor-free detectors.
By identifying and addressing the core issues that existed in the pseudo-labeling method on anchor-free detectors, our method can improve against the state-of-the-art methods. 
We further presented \proposedregmethod, a novel method that uses \textit{relative} Teacher/Student uncertainties to explicitly prevent the misleading regression pseudo-labels and select beneficial regression pseudo-labels in a boundary-wise manner. 
This enables the regression branch to benefit from the use of unlabeled images. 
In the experiment sections, we examine each method in three different SS-OD tasks and present consistent improvements. 
We also provide an extensive study to verify the effectiveness and generalization of our proposed~\proposedregmethod\ mechanism on both anchor-free and anchor-based detectors. 

Concerning negative societal impacts, we think it is essential to be aware that there exists the risk that object detection techniques (not just our method) are used in surveillance systems.
Also, as this line of works relies on low-labeled data for the model training, this aggravates the risk of data bias toward historically disadvantaged groups.

\section{Acknowledgments}
Yen-Cheng Liu and Zsolt Kira were partly supported by DARPA’s Learning with Less Labels (LwLL) program under agreement HR0011-18-S-0044, as part of their affiliation with Georgia Tech.


\newpage

{\small
\bibliographystyle{ieee_fullname}
\bibliography{reference}

\begin{thebibliography}{10}\itemsep=-1pt

\bibitem{berthelot2019mixmatch}
David Berthelot, Nicholas Carlini, Ian Goodfellow, Nicolas Papernot, Avital
  Oliver, and Colin~A Raffel.
\newblock Mixmatch: A holistic approach to semi-supervised learning.
\newblock In {\em Advances in Neural Information Processing Systems (NeurIPS)},
  pages 5049--5059, 2019.

\bibitem{cai2016unified}
Zhaowei Cai, Quanfu Fan, Rogerio~S Feris, and Nuno Vasconcelos.
\newblock A unified multi-scale deep convolutional neural network for fast
  object detection.
\newblock In {\em Proceedings of the European Conference on Computer Vision
  (ECCV)}, 2016.

\bibitem{chen2017learning}
Guobin Chen, Wongun Choi, Xiang Yu, Tony Han, and Manmohan Chandraker.
\newblock Learning efficient object detection models with knowledge
  distillation.
\newblock In {\em Advances in Neural Information Processing Systems (NeurIPS)},
  2017.

\bibitem{everingham2010pascal}
Mark Everingham, Luc Van~Gool, Christopher~KI Williams, John Winn, and Andrew
  Zisserman.
\newblock The pascal visual object classes (voc) challenge.
\newblock {\em International Journal of Computer Vision (IJCV)},
  88(2):303--338, 2010.

\bibitem{gao2019note}
Jiyang Gao, Jiang Wang, Shengyang Dai, Li-Jia Li, and Ram Nevatia.
\newblock Note-rcnn: Noise tolerant ensemble rcnn for semi-supervised object
  detection.
\newblock In {\em Proceedings of the IEEE International Conference on Computer
  Vision (ICCV)}, 2019.

\bibitem{guo2019mixup}
Hongyu Guo, Yongyi Mao, and Richong Zhang.
\newblock Mixup as locally linear out-of-manifold regularization.
\newblock In {\em Proceedings of the AAAI Conference on Artificial Intelligence
  (AAAI)}, volume~33, pages 3714--3722, 2019.

\bibitem{he2019bounding}
Yihui He, Chenchen Zhu, Jianren Wang, Marios Savvides, and Xiangyu Zhang.
\newblock Bounding box regression with uncertainty for accurate object
  detection.
\newblock In {\em Proceedings of the ieee/cvf conference on computer vision and
  pattern recognition}, pages 2888--2897, 2019.

\bibitem{hendrycks2020augmix}
Dan Hendrycks, Norman Mu, Ekin~D. Cubuk, Barret Zoph, Justin Gilmer, and Balaji
  Lakshminarayanan.
\newblock {AugMix}: A simple data processing method to improve robustness and
  uncertainty.
\newblock {\em Proceedings of the International Conference on Learning
  Representations (ICLR)}, 2020.

\bibitem{jeong2019consistency}
Jisoo Jeong, Seungeui Lee, Jeesoo Kim, and Nojun Kwak.
\newblock Consistency-based semi-supervised learning for object detection.
\newblock In {\em Advances in Neural Information Processing Systems (NeurIPS)},
  2019.

\bibitem{jiao2019survey}
Licheng Jiao, Fan Zhang, Fang Liu, Shuyuan Yang, Lingling Li, Zhixi Feng, and
  Rong Qu.
\newblock A survey of deep learning-based object detection.
\newblock {\em IEEE Access}, 7:128837--128868, 2019.

\bibitem{kong2020foveabox}
Tao Kong, Fuchun Sun, Huaping Liu, Yuning Jiang, Lei Li, and Jianbo Shi.
\newblock Foveabox: Beyound anchor-based object detection.
\newblock {\em IEEE Transactions on Image Processing}, 29:7389--7398, 2020.

\bibitem{laine2016temporal}
Samuli Laine and Timo Aila.
\newblock Temporal ensembling for semi-supervised learning.
\newblock In {\em Proceedings of the International Conference on Learning
  Representations (ICLR)}, 2017.

\bibitem{law2018cornernet}
Hei Law and Jia Deng.
\newblock Cornernet: Detecting objects as paired keypoints.
\newblock In {\em Proceedings of the European Conference on Computer Vision
  (ECCV)}, 2018.

\bibitem{lee2020localization}
Youngwan Lee, Joong-won Hwang, Hyung-Il Kim, Kimin Yun, and Joungyoul Park.
\newblock Localization uncertainty estimation for anchor-free object detection.
\newblock {\em arXiv preprint arXiv:2006.15607}, 2020.

\bibitem{li2020gflv2}
Xiang Li, Wenhai Wang, Xiaolin Hu, Jun Li, Jinhui Tang, and Jian Yang.
\newblock Generalized focal loss v2: Learning reliable localization quality
  estimation for dense object detection.
\newblock In {\em Proceedings of the IEEE Conference on Computer Vision and
  Pattern Recognition (CVPR)}, 2021.

\bibitem{li2020generalized}
Xiang Li, Wenhai Wang, Lijun Wu, Shuo Chen, Xiaolin Hu, Jun Li, Jinhui Tang,
  and Jian Yang.
\newblock Generalized focal loss: Learning qualified and distributed bounding
  boxes for dense object detection.
\newblock In {\em Advances in Neural Information Processing Systems (NeurIPS)},
  2020.

\bibitem{lin2017focal}
Tsung-Yi Lin, Priya Goyal, Ross Girshick, Kaiming He, and Piotr Doll{\'a}r.
\newblock Focal loss for dense object detection.
\newblock In {\em Proceedings of the IEEE international conference on computer
  vision (CVPR)}, pages 2980--2988, 2017.

\bibitem{lin2014microsoft}
Tsung-Yi Lin, Michael Maire, Serge Belongie, James Hays, Pietro Perona, Deva
  Ramanan, Piotr Doll{\'a}r, and C~Lawrence Zitnick.
\newblock Microsoft coco: Common objects in context.
\newblock In {\em Proceedings of the European Conference on Computer Vision
  (ECCV)}, 2014.

\bibitem{liu2016ssd}
Wei Liu, Dragomir Anguelov, Dumitru Erhan, Christian Szegedy, Scott Reed,
  Cheng-Yang Fu, and Alexander~C Berg.
\newblock Ssd: Single shot multibox detector.
\newblock In {\em European conference on computer vision (ECCV)}, pages 21--37.
  Springer, 2016.

\bibitem{liu2021unbiased}
Yen-Cheng Liu, Chih-Yao Ma, Zijian He, Chia-Wen Kuo, Kan Chen, Peizhao Zhang,
  Bichen Wu, Zsolt Kira, and Peter Vajda.
\newblock Unbiased teacher for semi-supervised object detection.
\newblock In {\em Proceedings of the International Conference on Learning
  Representations (ICLR)}, 2021.

\bibitem{ren2015faster}
Shaoqing Ren, Kaiming He, Ross Girshick, and Jian Sun.
\newblock Faster r-cnn: Towards real-time object detection with region proposal
  networks.
\newblock In {\em Advances in neural information processing systems (NeurIPS)},
  pages 91--99, 2015.

\bibitem{sajjadi2016regularization}
Mehdi Sajjadi, Mehran Javanmardi, and Tolga Tasdizen.
\newblock Regularization with stochastic transformations and perturbations for
  deep semi-supervised learning.
\newblock In {\em Advances in Neural Information Processing Systems (NeurIPS)},
  pages 1163--1171, 2016.

\bibitem{saputra2019distilling}
Muhamad Risqi~U Saputra, Pedro~PB de Gusmao, Yasin Almalioglu, Andrew Markham,
  and Niki Trigoni.
\newblock Distilling knowledge from a deep pose regressor network.
\newblock In {\em Proceedings of the IEEE International Conference on Computer
  Vision (ICCV)}, 2019.

\bibitem{shrivastava2016training}
Abhinav Shrivastava, Abhinav Gupta, and Ross Girshick.
\newblock Training region-based object detectors with online hard example
  mining.
\newblock In {\em Proceedings of the IEEE Conference on Computer Vision and
  Pattern Recognition (CVPR)}, 2016.

\bibitem{sohn2020fixmatch}
Kihyuk Sohn, David Berthelot, Chun-Liang Li, Zizhao Zhang, Nicholas Carlini,
  Ekin~D Cubuk, Alex Kurakin, Han Zhang, and Colin Raffel.
\newblock Fixmatch: Simplifying semi-supervised learning with consistency and
  confidence.
\newblock In {\em Advances in Neural Information Processing Systems (NeurIPS)},
  2020.

\bibitem{sohn2020simple}
Kihyuk Sohn, Zizhao Zhang, Chun-Liang Li, Han Zhang, Chen-Yu Lee, and Tomas
  Pfister.
\newblock A simple semi-supervised learning framework for object detection.
\newblock {\em arXiv preprint arXiv:2005.04757}, 2020.

\bibitem{tang2021humble}
Yihe Tang, Weifeng Chen, Yijun Luo, and Yuting Zhang.
\newblock Humble teachers teach better students for semi-supervised object
  detection.
\newblock In {\em Proceedings of the IEEE/CVF Conference on Computer Vision and
  Pattern Recognition}, pages 3132--3141, 2021.

\bibitem{tarvainen2017mean}
Antti Tarvainen and Harri Valpola.
\newblock Mean teachers are better role models: Weight-averaged consistency
  targets improve semi-supervised deep learning results.
\newblock In {\em Advances in neural information processing systems (NeurIPS)},
  pages 1195--1204, 2017.

\bibitem{tian2019fcos}
Zhi Tian, Chunhua Shen, Hao Chen, and Tong He.
\newblock Fcos: Fully convolutional one-stage object detection.
\newblock In {\em Proceedings of the IEEE International Conference on Computer
  Vision (ICCV)}, 2019.

\bibitem{wang2019region}
Jiaqi Wang, Kai Chen, Shuo Yang, Chen~Change Loy, and Dahua Lin.
\newblock Region proposal by guided anchoring.
\newblock In {\em Proceedings of the IEEE Conference on Computer Vision and
  Pattern Recognition (CVPR)}, 2019.

\bibitem{wu2019detectron2}
Yuxin Wu, Alexander Kirillov, Francisco Massa, Wan-Yen Lo, and Ross Girshick.
\newblock Detectron2.
\newblock \url{https://github.com/facebookresearch/detectron2}, 2019.

\bibitem{xie2017aggregated}
Saining Xie, Ross Girshick, Piotr Doll{\'a}r, Zhuowen Tu, and Kaiming He.
\newblock Aggregated residual transformations for deep neural networks.
\newblock In {\em Proceedings of the IEEE Conference on Computer Vision and
  Pattern Recognition (CVPR)}, 2017.

\bibitem{xu2021end}
Mengde Xu, Zheng Zhang, Han Hu, Jianfeng Wang, Lijuan Wang, Fangyun Wei, Xiang
  Bai, and Zicheng Liu.
\newblock End-to-end semi-supervised object detection with soft teacher.
\newblock {\em arXiv preprint arXiv:2106.09018}, 2021.

\bibitem{yang2021interactive}
Qize Yang, Xihan Wei, Biao Wang, Xian-Sheng Hua, and Lei Zhang.
\newblock Interactive self-training with mean teachers for semi-supervised
  object detection.
\newblock In {\em Proceedings of the IEEE/CVF Conference on Computer Vision and
  Pattern Recognition}, pages 5941--5950, 2021.

\bibitem{yun2019cutmix}
Sangdoo Yun, Dongyoon Han, Seong~Joon Oh, Sanghyuk Chun, Junsuk Choe, and
  Youngjoon Yoo.
\newblock Cutmix: Regularization strategy to train strong classifiers with
  localizable features.
\newblock In {\em Proceedings of the IEEE International Conference on Computer
  Vision (ICCV)}, pages 6023--6032, 2019.

\bibitem{zhang2018mixup}
Hongyi Zhang, Moustapha Cisse, Yann~N Dauphin, and David Lopez-Paz.
\newblock mixup: Beyond empirical risk minimization.
\newblock In {\em Proc. International Conference on Learning Representations
  (ICLR)}, 2018.

\bibitem{zhang2020bridging}
Shifeng Zhang, Cheng Chi, Yongqiang Yao, Zhen Lei, and Stan~Z Li.
\newblock Bridging the gap between anchor-based and anchor-free detection via
  adaptive training sample selection.
\newblock In {\em Proceedings of the IEEE Conference on Computer Vision and
  Pattern Recognition (CVPR)}, 2020.

\bibitem{zhou2021instant}
Qiang Zhou, Chaohui Yu, Zhibin Wang, Qi Qian, and Hao Li.
\newblock Instant-teaching: An end-to-end semi-supervised object detection
  framework.
\newblock In {\em Proceedings of the IEEE Conference on Computer Vision and
  Pattern Recognition (CVPR)}, 2021.

\bibitem{zhou2019objects}
Xingyi Zhou, Dequan Wang, and Philipp Kr{\"a}henb{\"u}hl.
\newblock Objects as points.
\newblock In {\em arXiv preprint arXiv:1904.07850}, 2019.

\bibitem{zhou2019bottom}
Xingyi Zhou, Jiacheng Zhuo, and Philipp Krahenbuhl.
\newblock Bottom-up object detection by grouping extreme and center points.
\newblock In {\em Proceedings of the IEEE Conference on Computer Vision and
  Pattern Recognition (CVPR)}, 2019.

\bibitem{zhu2019soft}
Chenchen Zhu, Fangyi Chen, Zhiqiang Shen, and Marios Savvides.
\newblock Soft anchor-point object detection.
\newblock In {\em Proceedings of the European Conference on Computer Vision
  (ECCV)}, 2020.

\bibitem{zhu2019feature}
Chenchen Zhu, Yihui He, and Marios Savvides.
\newblock Feature selective anchor-free module for single-shot object
  detection.
\newblock In {\em Proceedings of the IEEE Conference on Computer Vision and
  Pattern Recognition (CVPR)}, 2019.

\end{thebibliography}
}

\end{document}